\documentclass{article}

\PassOptionsToPackage{numbers, compress}{natbib}

\usepackage[preprint]{neurips_2024}

\usepackage{footmisc}
\usepackage[utf8]{inputenc} 
\usepackage[T1]{fontenc}    
\usepackage{hyperref}       
\usepackage{url}            
\usepackage{booktabs}       
\usepackage{amsfonts}       
\usepackage{nicefrac}       
\usepackage{microtype}      
\usepackage{xcolor}         

\usepackage{graphicx}
\usepackage{wrapfig}
\usepackage{subfigure}

\usepackage{amsmath,amssymb}

\usepackage{xspace}
\makeatletter
\DeclareRobustCommand\onedot{\futurelet\@let@token\@onedot}
\def\@onedot{\ifx\@let@token.\else.\null\fi\xspace}

\newcommand\figcaption{\def\@captype{figure}\caption}
\newcommand\tabcaption{\def\@captype{table}\caption}
\makeatother

\title{InstantCharacter: Personalize Any Characters with a Scalable Diffusion Transformer Framework}

\author{
    Jiale Tao$^{1}$, Yanbing Zhang$^{1}$, Qixun Wang$^{12}$\footnotemark[2]\ , \\ 
    \textbf{Yiji Cheng}$^{1}$, \textbf{Haofan Wang}$^{2}$, \textbf{Xu Bai}$^{2}$, \textbf{Zhengguang Zhou}$^{12}$, \textbf{Ruihuang Li}$^{1}$, \textbf{Linqing Wang}$^{12}$, \\
    \textbf{Chunyu Wang}$^{1}$, \textbf{Qin Lin}$^{1}$, \textbf{Qinglin Lu}$^{1}$\footnotemark[1]\\
    $^{1}$Hunyuan, Tencent \ 
    $^{2}$InstantX Team\\
    \footnotemark[2]\ \ Tech Lead, \footnotemark[1]\ \ Corresponding Author
}

\begin{document}

\maketitle

\begin{abstract}
Current learning-based subject customization approaches, predominantly relying on U-Net architectures, suffer from limited generalization ability and compromised image quality. Meanwhile, optimization-based methods require subject-specific fine-tuning, which inevitably degrades textual controllability. To address these challenges, we propose InstantCharacter—a scalable framework for character customization built upon a foundation diffusion transformer. InstantCharacter demonstrates three fundamental advantages: first, it achieves open-domain personalization across diverse character appearances, poses, and styles while maintaining high-fidelity results. Second, the framework introduces a scalable adapter with stacked transformer encoders, which effectively processes open-domain character features and seamlessly interacts with the latent space of modern diffusion transformers. Third, to effectively train the framework, we construct a large-scale character dataset containing 10-million-level samples. The dataset is systematically organized into paired (multi-view character) and unpaired (text-image combinations) subsets. This dual-data structure enables simultaneous optimization of identity consistency and textual editability through distinct learning pathways. Qualitative experiments demonstrate InstantCharacter’s advanced capabilities in generating high-fidelity, text-controllable, and character-consistent images, setting a new benchmark for character-driven image generation. Our source code is available at https://github.com/Tencent/InstantCharacter. 

\end{abstract}

\begin{figure*}[ht]
  \centering
  \includegraphics[width=1.0\linewidth]{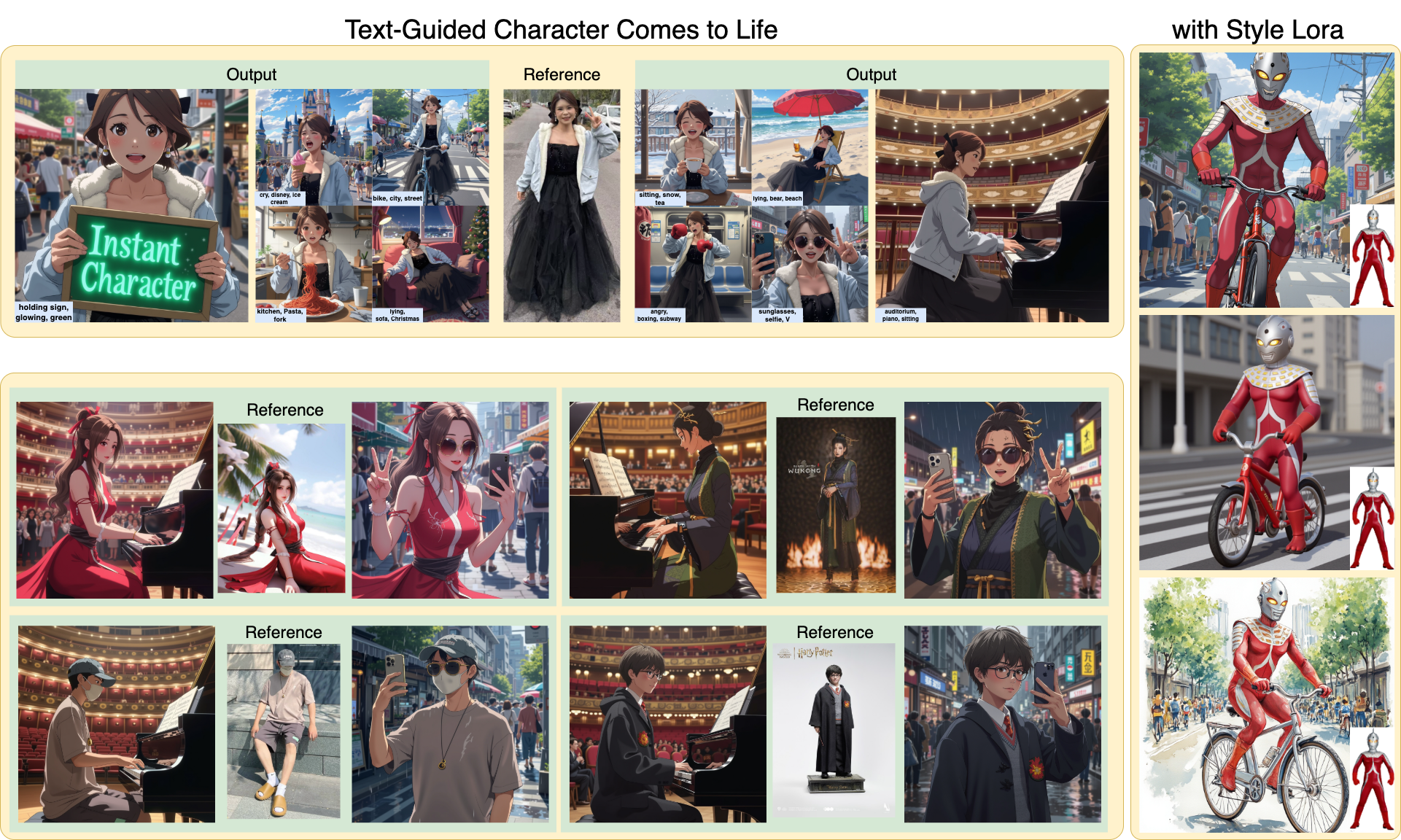}
  \caption{Open-domain character personalization with InstantCharacter.}
  \label{fig:intro}
\end{figure*}

\section{Introduction}
\label{sec:intro}

Character-driven image generation aims to create images that incorporates the user-defined character image and text prompts, playing a crucial role in various creative endeavors such as storytelling illustration, comic creation, game character design, and more. These capabilities enable a wide range of applications in entertainment, film production, e-commerce advertising, and beyond. Recent advancements in generative diffusion transformers have demonstrated unprecedented capabilities in synthesizing high-fidelity images from textual descriptions. Nevertheless, the potential of these state-of-the-art models for personalized image generation remains underexplored, especially in the context of creating character-driven visual narratives that embody human-like attributes.

Current methodologies for generating consistent images of specified subjects primarily rely on tuning- or adapter-based approaches. Adapter-based approaches~\cite{li2023blip,ye2023ip,mou2024t2i} extract visual features through a subject encoder and integrate them into the image noise space via cross-attention mechanism. While these techniques achieve certain subject consistency and text controllability on UNet-based models, they struggle to personalize open-domain characters with diverse identities, poses, and styles. Although effective for customizing open-domain characters, tuning-based approaches~\cite{ruiz2023dreambooth} require fine-tuning the model to reconstruct subject images, leading to long customization time and limited text controllability. Moreover, inference-time fine-tuning becomes computationally prohibitive for modern diffusion transformers with billions of parameters.

Compared to traditional UNet-based architectures~\cite{rombach2022sd,podell2023sdxl}, modern Diffusion Transformers (DiTs)~\cite{esser2024sd3,flux} exhibit powerful generative priors and offer unparalleled flexibility and capacity. However, fully unleashing their potential is non-trivial, as it requires a robust adapter network compatible with the framework to ensure alignment between character-specific features and vast generative latent space. In addition, training such an adapter necessitates adequate training data and effective training strategies. We observe that directly applying traditional adapters to large-scale DiTs often fails, as these adapters are primarily designed for UNet architectures and cannot scale effectively to models with billions of parameters, such as Flux~\cite{flux} with 12 billion parameters.

To achieve generalized character personalization without compromising inference-time efficiency and textual editability, we propose InstantCharacter, a scalable diffusion transformer framework designed for character-driven image generation. InstantCharacter offers three key advantages: 1.\textbf{Generalizability.} It can flexibly personalize any character with different appearances, actions, and styles, ranging from photorealistic portraits to anime game assets. 2.\textbf{Scalability.} We develop a scalable adapter that can effectively integrate multi-stage character features and interact with the latent space of modern DiTs. 3.\textbf{Versatility.} To enable efficient training, we collect a versatile 10-million-level character dataset, which contains paired (multi-view character) and unpaired (text-image combinations) subsets. Accordingly, we propose an efficient three-stage training strategy to accommodate heterogeneous data samples. Specifically, we decouple character consistency (unpaired data), textual controllability (paired data), and image fidelity (high-resolution data) to prevent mutual interference between high-fidelity identity maintenance and prompt-guided character manipulations.

We implement InstantCharacter based on the powerful FLUX1.0-dev model. Qualitative comparisons with previous work demonstrate InstantCharacter’s advanced capabilities in generating high-fidelity, text-controllable, and character-consistent images.


\section{Related Work}
\label{sec:related}
\noindent\textbf{T2I diffusion models.} Recent advances~\cite{esser2024sd3,podell2023sdxl,rombach2022sd} in text-to-image generation have witnessed a paradigm shift from traditional U-Net architectures~\cite{rombach2022sd} to more powerful diffusion transformers~\cite{esser2024sd3} (DiTs). While early diffusion models such as stable diffusion (SD) demonstrated remarkable image synthesis capabilities, modern DiT-based systems like SD3~\cite{esser2024sd3} and FLUX.1~\cite{flux} have set new benchmarks in generation quality through their transformer-based architectures and advanced techniques like rectified flows. This architectural evolution presents both opportunities and challenges for character-centric applications, while DiTs offer superior generation capacity, their adaptation for identity-preserving tasks remains largely underexplored. Our work bridges this critical gap by developing the first DiT-based framework specifically optimized for character customization.

\noindent\textbf{Personalized character generation.} Recent advances in personalized image generation have evolved from tuning-based to adapter-based approaches. Early methods~\cite{ruiz2023dreambooth,chefer2023attend,feng2025personalize,kumari2023multi,textinversion} relied on fine-tuning the entire diffusion model for each new subject, which was computationally expensive and suffered from poor generalization due to limited training data. To address these issues, recent works~\cite{ye2023ip,li2023blip,mou2024t2i,wang2024instantid,li2024photomaker,huang2024realcustom,wang2024characterfactory,mao2024realcustom++} introduced adapter-based techniques that avoid test-time fine-tuning. For instance, IP-Adapter~\cite{ye2023ip} employs a clip image encoder to extract subject features and injects them into a frozen diffusion model via cross-attention, enabling efficient personalization. However, these adapter-based methods are predominantly built upon UNet-based architectures with restricted capacity, causing them to struggle in effectively scaling and often produce low-fidelity outputs and limited generalization across diverse character poses and styles. In contrast, our work introduces a scalable diffusion transformer framework that overcomes these limitations, achieving superior open-domain generalizability, image fidelity, and text controllability compared to UNet-based alternatives.

\section{Methods}
\label{sec:formatting}

\begin{figure*}[ht]
  \centering
  \includegraphics[width=1.0\linewidth]{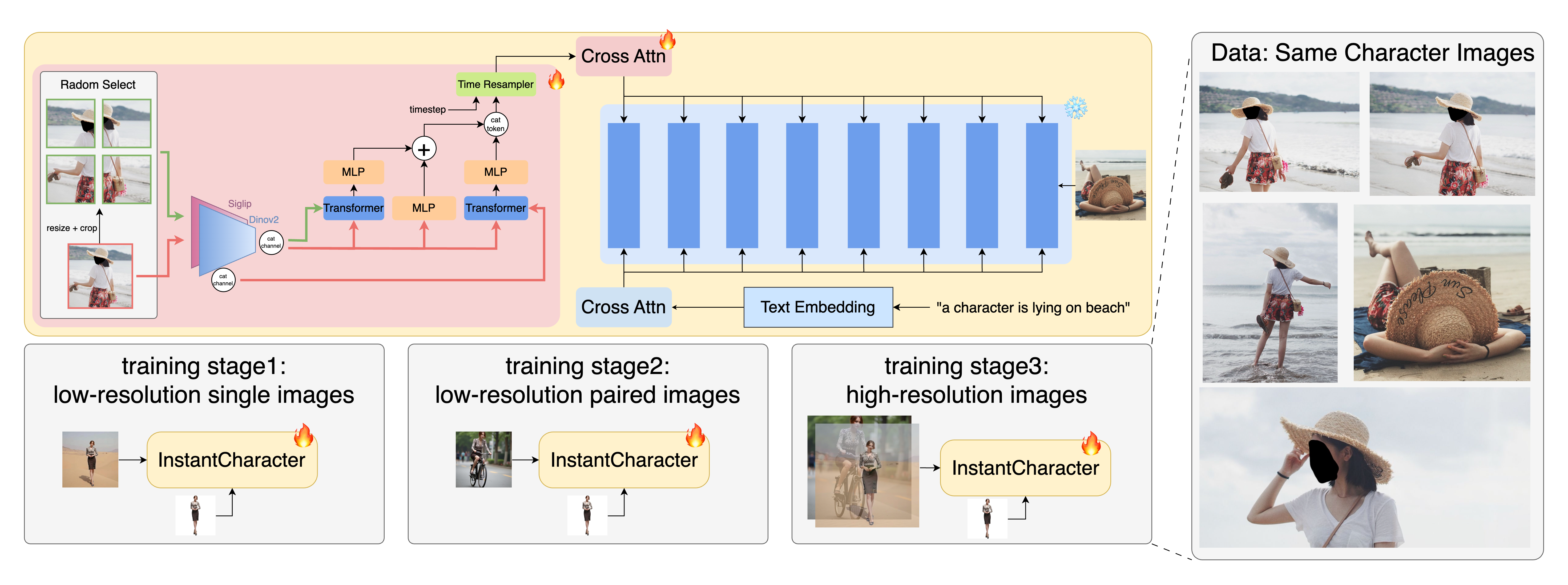}
  \caption{Our framework seamlessly integrates a scalable adapter with a pretrained DiT model. The adapter consists of multiple stacked transformer encoders that incrementally refine character representations, enabling effective interaction with the latent space of the DiT. The training process employs a three-stage progressive strategy, beginning with unpaired low-resolution pretraining and culminating in paired high-resolution fine-tuning.}
  \label{fig:pipeline}
\end{figure*}

Modern DiTs~\cite{esser2024sd3,flux} have demonstrated unprecedented fidelity and capacity compared to traditional UNet-based architectures, offering a more robust foundation for generation and editing tasks. 
Building upon these advances, we present InstantCharacter, a novel framework that extends DiT for generalizable and high-fidelity character-driven image generation. As illustrated in Figure~\ref{fig:pipeline}, InstantCharacter's architecture centers around two key innovations. First, a scalable adapter module is developed to effectively parse character features and seamlessly interact with DiTs latent space. Second, a progressive three-stage training strategy is designed to adapt to our collected versatile dataset, enabling separate training for character consistency and text editability. By synergistically combining flexible adapter design and phased learning strategy, we enhance the general character customization capability while maximizing the preservation of the generative priors of the base DiT model. In the following sections, we will detail the adapter's architecture and elaborate on our progressive training strategy.

\subsection{The scalable adapter design}

Traditional customization adapters, such as IPAdapter~\cite{ye2023ip} or ReferenceNet~\cite{song2024moma}, often fail in the DiT architectures because they are specifically designed for U-Net based models and lack scalability. To better adapt to DiT models, we propose a scalable full-transformer adapter that serves as a crucial link between conditioning character images and the latent generative space of the base model. The full-transformer structure enables scalability by increasing layer depth and hidden feature sizes. This adapter consists of three encoder blocks, as detailed below.

\noindent\textbf{General vision encoders.} We first leverage pre-trained large vision foundation encoders to extract general character features, benefiting from their open-domain recognition abilities. Previous methods~\cite{ye2023ip,li2024photomaker} typically rely on CLIP~\cite{radford2021learning} for its aligned visual and textual features. However, while CLIP is capable of capturing abstract semantic information, it tends to lose detailed texture information, which is crucial for maintaining character consistency. To this end, we replace CLIP with SigLIP~\cite{zhai2023sigmoid}, which excels in capturing finer-grained character information. In addition, we introduce DINOv2~\cite{oquab2023dinov2} as another image encoder to enhance the robustness of features, reducing the loss of features caused by background or other interfering factors. Finally, we integrate DINOv2 and SigLIP features via channel-wise concatenation, resulting in a more comprehensive representation of open-domain characters.

\noindent\textbf{Intermediate encoders.} Since SigLIP and DINOv2 are pre-trained and inferred at a relatively low resolution of 384, the raw output of general vision encoders may lose fine-grained features when processing high-resolution character images. To mitigate this issue, we employ a dual-stream feature fusion strategy to explore low-level and region-level features, respectively. First, we directly extract \textbf{low-level features} from the shallow layers of the general vision encoders, capturing details that are often lost in higher layers. Second, we divide the reference image into multiple non-overlapping patches and feed each patch into the vision encoder to obtain \textbf{region-level} features. Then these two distinct feature streams undergo hierarchical integration through dedicated intermediate transformer encoders. Specifically, each feature pathway is independently processed by a separate transformer encoder to integrate with high-level semantic features. Subsequently, the refined feature embeddings from both pathways are concatenated along the token dimension, establishing a comprehensive fused representation that captures multi-level complementary information.

\noindent\textbf{Projection head.}  Finally, the refined character features are projected into the denoising space via a projection head and interact with the latent noise. We implement this through a timestep-aware Q-former~\cite{ye2023ip} that processes intermediate encoder outputs as key-value pairs while dynamically updating a set of learnable queries through attention mechanisms. The transformed query features are then injected into the denoising space via learnable cross-attention layers. Finally, the adapter enables faithful identity preservation and flexible adaptation to complex text-driven modifications.

\subsection{Training strategies}

To enable effective training of the framework, we first curate a high-quality dataset of 10 million images containing diverse full-body humans/characters, including both unpaired images for learning robust character consistency and paired sets for achieving precise text-to-image alignment. 

Our training regimen is meticulously designed to optimize character consistency, text controllability, and visual fidelity. To achieve character consistency, we first train with unpaired data, where the character image is incorporated as reference guidance to reconstruct itself and preserve structural consistency. We discovered that using a resolution of 512 is significantly more efficient than 1024. 

In the second phase, we continue training at a low resolution (512) but switch to paired training data. By taking the character image as input, we aim to generate images of the character in different actions, poses, and styles within a new scene based on a given textual description. This training stage efficiently eliminates the copy-paste effect and enhances text controllability, ensuring that the generated images accurately follow the textual condition.

The final phase involves high-resolution joint training using both paired and non-paired images. We found that a limited number of high-resolution training iterations can substantially improve the visual quality and texture of the images. This stage leverages high-quality images to achieve high-fidelity and textually controlled character images.

\section{Experiments}

\begin{figure*}[ht]
  \centering
  \includegraphics[width=1.0\linewidth]{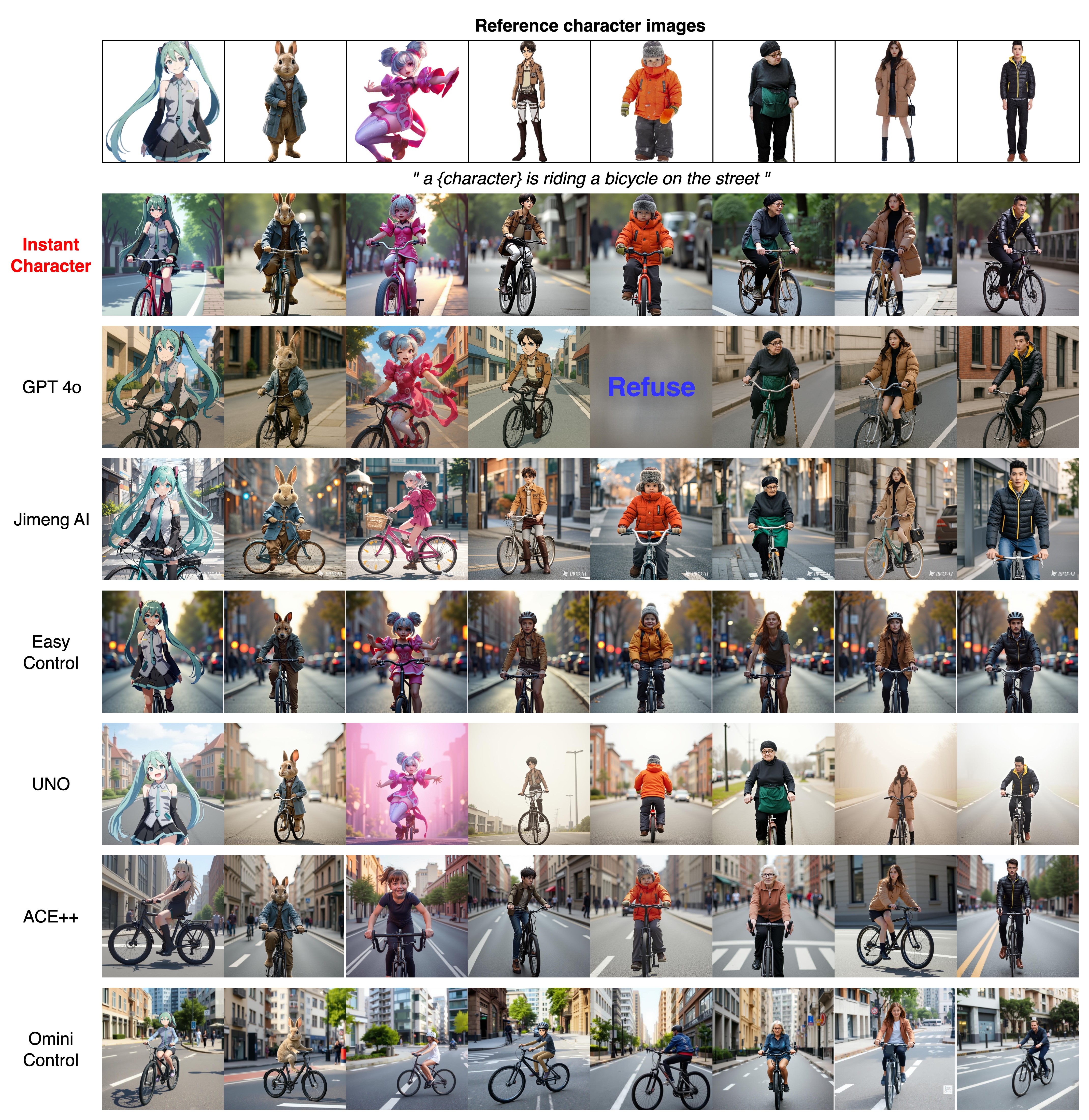}
  \caption{Qualitative comparison on character personalization. Our method generally demonstrates the best image fidelity and character consistency while maintaining the desirable textual controllability.}
  \label{fig:compare}
  \vspace{-0.3cm}
\end{figure*}

\begin{figure*}[ht]
  \centering
  \includegraphics[width=1.0\linewidth]{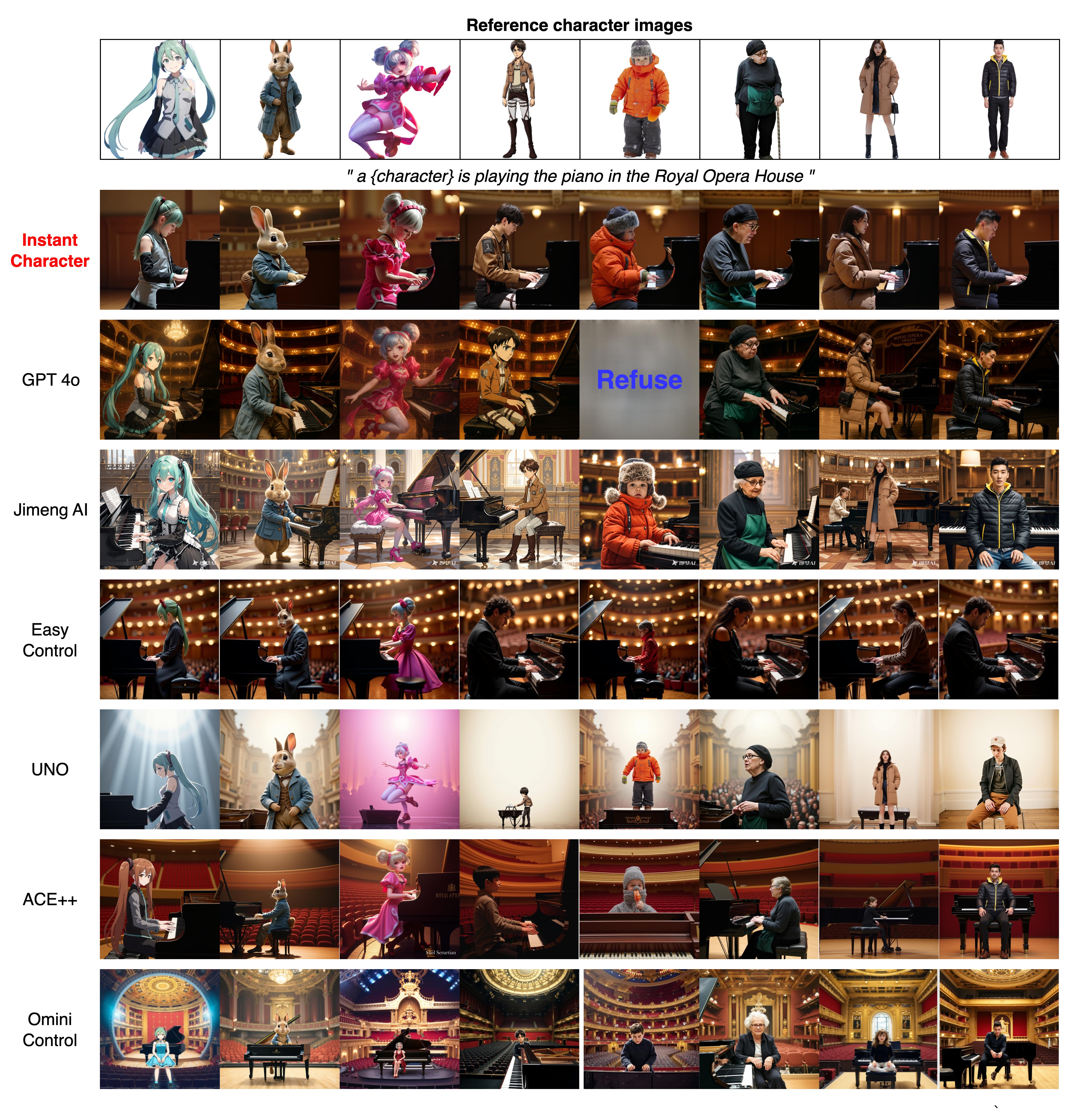}
  \caption{Qualitative comparison on character personalization. Our method generally demonstrates the best image fidelity and character consistency while maintaining the desirable textual controllability.}
  \label{fig:compare2}
  \vspace{-0.3cm}
\end{figure*}

\begin{figure*}[ht]
  \centering
  \includegraphics[width=1.0\linewidth]{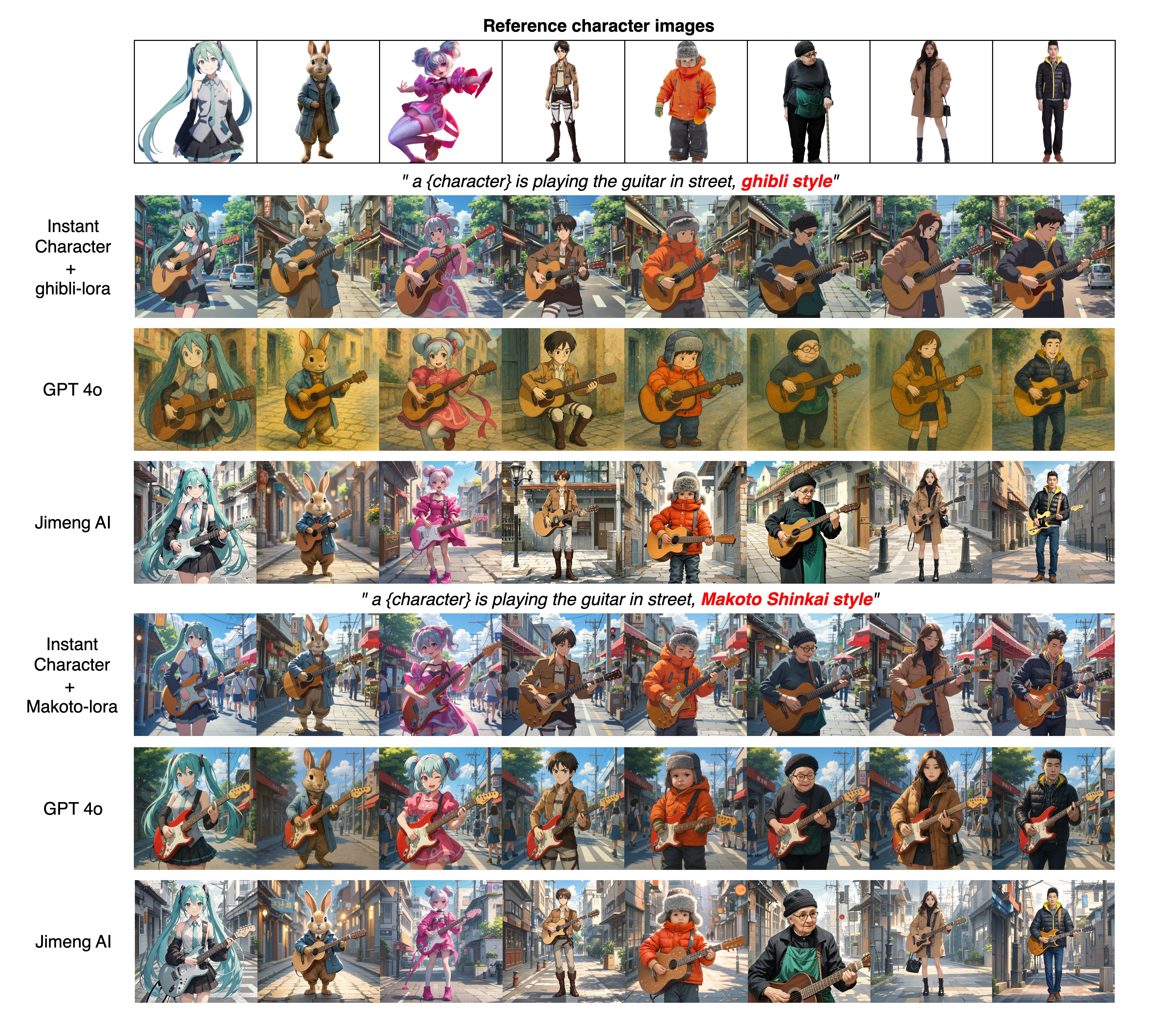}
  \caption{Qualitative comparison on character personalization with different styles.}
  \label{fig:compare_style}
  \vspace{-0.3cm}
\end{figure*}

\begin{figure*}[ht]
  \centering
  \includegraphics[width=1.0\linewidth]{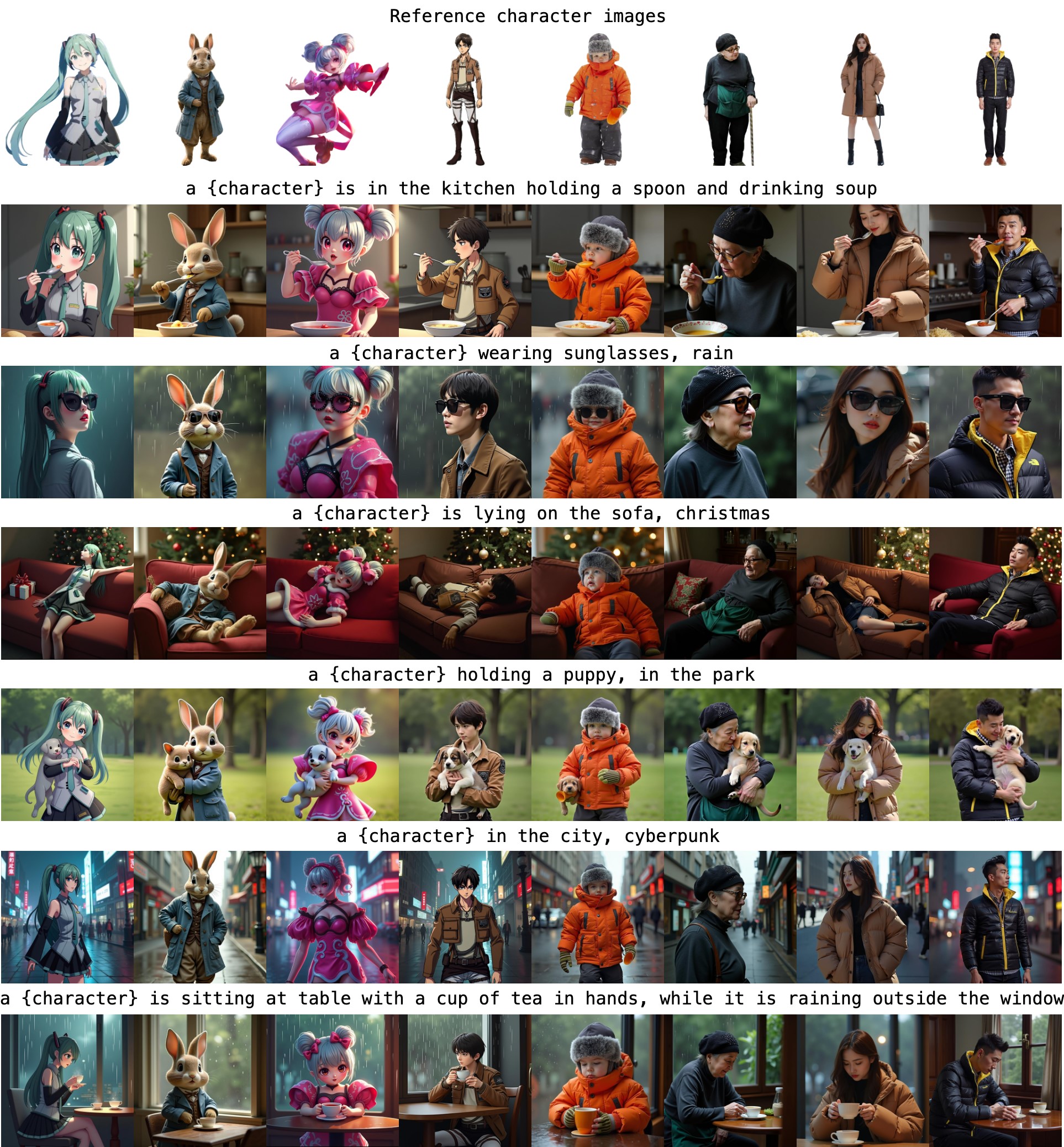}
  \caption{More qualitative results of InstantCharacter.}
  \label{fig:qualitative}
  \vspace{-0.3cm}
\end{figure*}

\noindent\textbf{Qualitative results.}  We conduct qualitative comparisons against state-of-the-art FLUX-based approaches: OminiControl~\cite{tan2024ominicontrol}, EasyControl~\cite{zhang2025easycontrol}, ACE+~\cite{mao2025ace++}, and UNO~\cite{UNO}; and the large multi-modality model GPT4o~\cite{4o}. For evaluation, we collect a set of open-domain character images not present in the training data. As shown in Fig.~\ref{fig:compare} and Fig.~\ref{fig:compare2}, our analysis demonstrates that while existing methods show limitations: OminiControl and EasyControl fail to preserve character identity features, and ACE++ only maintains partial features in simple scenarios while struggling with action-oriented prompts. UNO overly preserves consistency, which reduces the editability of actions and backgrounds. It is notable that our method achieves comparable results with GPT4o, which is the current SoTA method but it is not open source. In contrast, InstantCharacter consistently performs the best. Specifically, InstantCharacter achieves superior character detail preservation with high fidelity while maintaining precise text controllability, even for complex action prompts. These qualitative advantages are further supported by quantitative measurements shown in Fig.~\ref{fig:qualitative}.

\noindent\textbf{Personalization with different styles.}
Our framework can also achieve flexible character stylization by introducing different style loras. As shown in Fig.~\ref{fig:compare_style}, our method can switch between Ghibli and Makoto styles without compromising character consistency and textual editability. However, it is difficult for Jimeng and GPT4o to preserve the styles flexibly.

\section{Conclusion}
We present InstantCharacter, an innovative diffusion transformer framework that significantly advances character-driven image generation. Our solution delivers three fundamental advantages: first, it achieves unprecedented open-domain personalization across diverse character appearances, poses, and styles while preserving high-fidelity quality; second, it develops a scalable adapter architecture that effectively processes character features and interacts with diffusion transformers' latent space; third, it establishes an effective three-stage 
 training methodology combining a massive 10-million-scale dataset to simultaneously optimize character consistency and textual control. Qualitative results validate InstantCharacter's superior performance in generating high-fidelity, character-consistent, and text-controllable images. More broadly, our work offers insights for adapting foundation diffusion transformers to specialized generation tasks, potentially inspiring new developments in controllable visual synthesis.

\bibliographystyle{splncs04}
\bibliography{neurips_2024}

\begin{thebibliography}{10}
\providecommand{\url}[1]{\texttt{#1}}
\providecommand{\urlprefix}{URL }
\providecommand{\doi}[1]{https://doi.org/#1}

\bibitem{chefer2023attend}
Chefer, H., Alaluf, Y., Vinker, Y., Wolf, L., Cohen-Or, D.: Attend-and-excite: Attention-based semantic guidance for text-to-image diffusion models. ACM transactions on Graphics (TOG)  \textbf{42}(4),  1--10 (2023)

\bibitem{esser2024sd3}
Esser, P., Kulal, S., Blattmann, A., Entezari, R., M{\"u}ller, J., Saini, H., Levi, Y., Lorenz, D., Sauer, A., Boesel, F., et~al.: Scaling rectified flow transformers for high-resolution image synthesis. In: Forty-first international conference on machine learning (2024)

\bibitem{feng2025personalize}
Feng, H., Huang, Z., Li, L., Lv, H., Sheng, L.: Personalize anything for free with diffusion transformer. arXiv preprint arXiv:2503.12590  (2025)

\bibitem{textinversion}
Gal, R., Alaluf, Y., Atzmon, Y., Patashnik, O., Bermano, A.H., Chechik, G., Cohen-Or, D.: An image is worth one word: Personalizing text-to-image generation using textual inversion. arXiv preprint arXiv:2208.01618  (2022)

\bibitem{huang2024realcustom}
Huang, M., Mao, Z., Liu, M., He, Q., Zhang, Y.: Realcustom: narrowing real text word for real-time open-domain text-to-image customization. In: Proceedings of the IEEE/CVF Conference on Computer Vision and Pattern Recognition. pp. 7476--7485 (2024)

\bibitem{4o}
Hurst, A., Lerer, A., Goucher, A.P., Perelman, A., Ramesh, A., Clark, A., Ostrow, A., Welihinda, A., Hayes, A., Radford, A., et~al.: Gpt-4o system card. arXiv preprint arXiv:2410.21276  (2024)

\bibitem{kumari2023multi}
Kumari, N., Zhang, B., Zhang, R., Shechtman, E., Zhu, J.Y.: Multi-concept customization of text-to-image diffusion. In: Proceedings of the IEEE/CVF conference on computer vision and pattern recognition. pp. 1931--1941 (2023)

\bibitem{flux}
Labs, B.F.: Flux: Official inference repository for flux.1 models  (2024)

\bibitem{li2023blip}
Li, D., Li, J., Hoi, S.: Blip-diffusion: Pre-trained subject representation for controllable text-to-image generation and editing. Advances in Neural Information Processing Systems  \textbf{36},  30146--30166 (2023)

\bibitem{li2024photomaker}
Li, Z., Cao, M., Wang, X., Qi, Z., Cheng, M.M., Shan, Y.: Photomaker: Customizing realistic human photos via stacked id embedding. In: Proceedings of the IEEE/CVF conference on computer vision and pattern recognition. pp. 8640--8650 (2024)

\bibitem{mao2025ace++}
Mao, C., Zhang, J., Pan, Y., Jiang, Z., Han, Z., Liu, Y., Zhou, J.: Ace++: Instruction-based image creation and editing via context-aware content filling. arXiv preprint arXiv:2501.02487  (2025)

\bibitem{mao2024realcustom++}
Mao, Z., Huang, M., Ding, F., Liu, M., He, Q., Zhang, Y.: Realcustom++: Representing images as real-word for real-time customization. arXiv preprint arXiv:2408.09744  (2024)

\bibitem{mou2024t2i}
Mou, C., Wang, X., Xie, L., Wu, Y., Zhang, J., Qi, Z., Shan, Y.: T2i-adapter: Learning adapters to dig out more controllable ability for text-to-image diffusion models. In: Proceedings of the AAAI conference on artificial intelligence. vol.~38, pp. 4296--4304 (2024)

\bibitem{oquab2023dinov2}
Oquab, M., Darcet, T., Moutakanni, T., Vo, H., Szafraniec, M., Khalidov, V., Fernandez, P., Haziza, D., Massa, F., El-Nouby, A., et~al.: Dinov2: Learning robust visual features without supervision. arXiv preprint arXiv:2304.07193  (2023)

\bibitem{podell2023sdxl}
Podell, D., English, Z., Lacey, K., Blattmann, A., Dockhorn, T., M{\"u}ller, J., Penna, J., Rombach, R.: Sdxl: Improving latent diffusion models for high-resolution image synthesis. arXiv preprint arXiv:2307.01952  (2023)

\bibitem{radford2021learning}
Radford, A., Kim, J.W., Hallacy, C., Ramesh, A., Goh, G., Agarwal, S., Sastry, G., Askell, A., Mishkin, P., Clark, J., et~al.: Learning transferable visual models from natural language supervision. In: International conference on machine learning. pp. 8748--8763. PmLR (2021)

\bibitem{rombach2022sd}
Rombach, R., Blattmann, A., Lorenz, D., Esser, P., Ommer, B.: High-resolution image synthesis with latent diffusion models. In: Proceedings of the IEEE/CVF conference on computer vision and pattern recognition. pp. 10684--10695 (2022)

\bibitem{ruiz2023dreambooth}
Ruiz, N., Li, Y., Jampani, V., Pritch, Y., Rubinstein, M., Aberman, K.: Dreambooth: Fine tuning text-to-image diffusion models for subject-driven generation. In: Proceedings of the IEEE/CVF conference on computer vision and pattern recognition. pp. 22500--22510 (2023)

\bibitem{song2024moma}
Song, K., Zhu, Y., Liu, B., Yan, Q., Elgammal, A., Yang, X.: Moma: Multimodal llm adapter for fast personalized image generation. In: European Conference on Computer Vision. pp. 117--132. Springer (2024)

\bibitem{tan2024ominicontrol}
Tan, Z., Liu, S., Yang, X., Xue, Q., Wang, X.: Ominicontrol: Minimal and universal control for diffusion transformer. arXiv preprint arXiv:2411.15098  \textbf{3} (2024)

\bibitem{wang2024characterfactory}
Wang, Q., Li, B., Li, X., Cao, B., Ma, L., Lu, H., Jia, X.: Characterfactory: Sampling consistent characters with gans for diffusion models. arXiv preprint arXiv:2404.15677  (2024)

\bibitem{wang2024instantid}
Wang, Q., Bai, X., Wang, H., Qin, Z., Chen, A., Li, H., Tang, X., Hu, Y.: Instantid: Zero-shot identity-preserving generation in seconds. arXiv preprint arXiv:2401.07519  (2024)

\bibitem{UNO}
Wu, S., Huang, M., Wu, W., Cheng, Y., Ding, F., He, Q.: Less-to-more generalization: Unlocking more controllability by in-context generation. arXiv preprint arXiv:2504.02160  (2025)

\bibitem{ye2023ip}
Ye, H., Zhang, J., Liu, S., Han, X., Yang, W.: Ip-adapter: Text compatible image prompt adapter for text-to-image diffusion models. arXiv preprint arXiv:2308.06721  (2023)

\bibitem{zhai2023sigmoid}
Zhai, X., Mustafa, B., Kolesnikov, A., Beyer, L.: Sigmoid loss for language image pre-training. In: Proceedings of the IEEE/CVF international conference on computer vision. pp. 11975--11986 (2023)

\bibitem{zhang2025easycontrol}
Zhang, Y., Yuan, Y., Song, Y., Wang, H., Liu, J.: Easycontrol: Adding efficient and flexible control for diffusion transformer. arXiv preprint arXiv:2503.07027  (2025)

\end{thebibliography}

\end{document}